\documentclass[acmtog]{acmart}
\usepackage{booktabs}
\setcopyright{none}

\renewcommand\footnotetextcopyrightpermission[1]{}
\begin{document}

% Title. 
\title{ 3D Scene Parsing via Class-Wise Adaptation} 

% Authors.
\author{Daichi Ono}
\affiliation{%
  \institution{Sony Interactive Entertainment Inc.}}
\author{Hiroyuki Yabe}
\affiliation{%
  \institution{Sony Interactive Entertainment Inc.}}
\author{Tsutomu Horikawa}
\affiliation{%
  \institution{Sony Interactive Entertainment Inc.}}
\authorsaddresses{}

% Abstract
\begin{abstract}
We propose the method that uses only computer graphics datasets to parse the real world 3D scenes.
3D scene parsing based on semantic segmentation is required to implement the categorical interaction in the virtual world.
Convolutional Neural Networks (CNNs) have recently shown state-of-the-art performance on computer vision tasks including semantic segmentation. 
However, collecting and annotating a huge amount of data are needed to train CNNs.
Especially in the case of semantic segmentation, annotating pixel by pixel takes a significant amount of time and often makes mistakes.
In contrast, computer graphics can generate a lot of accurate annotated data and easily scale up by changing camera positions, textures and lights.
Despite these advantages, models trained on computer graphics datasets cannot perform well on real data, which is known as the domain shift.
To address this issue, we first present that depth modal and synthetic noise are effective to reduce the domain shift.
Then, we develop the class-wise adaptation which obtains domain invariant features of CNNs.
To reduce the domain shift, we create computer graphics rooms with a lot of props, and provide photo-realistic rendered images.
We also demonstrate the application which is combined semantic segmentation with Simultaneous Localization and Mapping (SLAM).
Our application performs accurate 3D scene parsing in real-time on an actual room.
\end{abstract}

% CCS
\begin{CCSXML}
<ccs2012>
<concept>
<concept_id>10010147.10010178.10010224.10010225.10010227</concept_id>
<concept_desc>Computing methodologies~Scene understanding</concept_desc>
<concept_significance>500</concept_significance>
</concept>
<concept>
<concept_id>10010147.10010178.10010224.10010245.10010254</concept_id>
<concept_desc>Computing methodologies~Reconstruction</concept_desc>
<concept_significance>500</concept_significance>
</concept>
<concept>
<concept_id>10010147.10010257.10010258.10010262.10010277</concept_id>
<concept_desc>Computing methodologies~Transfer learning</concept_desc>
<concept_significance>500</concept_significance>
</concept>
</ccs2012>
\end{CCSXML}

\ccsdesc[500]{Computing methodologies~Scene understanding}
\ccsdesc[500]{Computing methodologies~Reconstruction}
\ccsdesc[500]{Computing methodologies~Transfer learning}

% Keywords
\keywords{domain adaptation, scene parsing, neural networks, computer graphics, mixed-reality}

\begin{teaserfigure}
  \includegraphics[keepaspectratio=true, width=\textwidth]{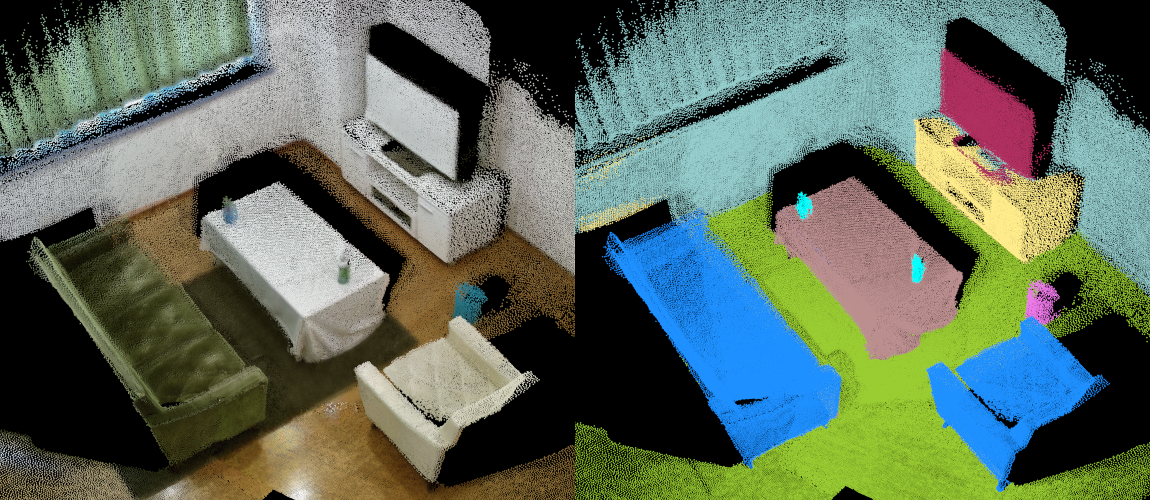}
  \caption{Our 3D scene parsing result. 
  The left shows actual rgb image and the right shows the parsing result. 
  The scene is captured as point cloud format, and each color of the parsing result shows each object class.}
  \label{fig:teaser}
\end{teaserfigure}

\maketitle
\thispagestyle{empty}

% Head 1
\section{Introduction}
3D scene parsing is a task of recognizing all objects densely as 3D in the real world scene.
3D scene parsing is required to implement the categorical interaction with real objects and the virtual objects.
For instance, in a mixed-reality application, walls are removed and chairs are replaced with computer graphics chairs.
There are several methods for 3D scene parsing. 
In this work, we use the method which is combined semantic segmentation with 3D reconstruction via visual SLAM.

Semantic segmentation is one kind of visual recognition task that assigns a categorical label to each pixel. 
It provides a wide range of applications such as autonomous driving, robotic navigation and mixed-reality entertainment. 
Recently, a lot of works that are based on Convolutional Neural Networks (CNNs) are improving performance of image recognition tasks. 
Since the performance of CNNs depends on the diversity of data, a significant amount of annotated data is required to further improve performance.
The main challenge is collecting and annotating data. 
In the case of indoor semantic segmentation, we have to collect millions of various pictures of rooms. 
In addition, annotating pixel by pixel is time consuming, as it takes more than 1 hour for a single image. 
Several approaches that use computer graphics datasets for training have been proposed to tackle those issues. 
If we use computer graphics datasets, we can easily increase pixel by pixel annotated images by changing camera positions, textures and lights, or replacing objects. 

However, due to a gap between computer graphics domain data and real domain data, models trained on computer graphics data cannot perform well on real data, 
which is known as the domain shift problem.
As semi-supervised methods, pre-training with computer graphics data and fine-tuning with real data, or mixing computer graphics data into
training mini-batch are proposed. 
However, these methods require certain amount of supervised real data to obtain better performance.
As unsupervised methods, for instance, the method that obtains domain invariant features of CNNs by adversarial training is proposed.
They have shown successful results on classification tasks which do not include object localization.
In contrast, on semantic segmentation, it is quite difficult to obtain the same performance as classification. 
It is assumed that the number of dimensions of features for adversarial training is too large.

First, in terms of input data, we present the effect of adding depth modal and synthetic noise to reduce the domain shift.
Then, in terms of features of CNNs, we propose the class-wise adaptation that is based on unsupervised adversarial training. 
By the class-wise adaptation, we can reduce the number of dimensions of features for adversarial training and simplify the task.

Since no annotated data are required, our method will overcome the domain shifts between computer graphics data and each room data,
and implement the high accurate mixed-reality application on each room.
As one example, we demonstrate the real world 3D scene parsing which is combined semantic segmentation with SLAM.

Our main contribution is that we propose a method for real world parsing using CNNs that are trained only computer graphics data. 
In establishing the method, we develop the class-wise adaptation that obtains domain invariant features from CNNs,
and present the effect of depth modal and synthetic noise for input data.
We also provide computer graphics rooms with a lot of props and photo-realistic rendered images to train the CNNs.

% Head 1
\section{Related Work}
In this section, we first review CNNs methods and datasets for semantic segmentation.
Next, we review computer graphics datasets creations for training data. 
Finally, we review domain adaptation methods which tackle domain shifts between training data and testing data.

\paragraph{Semantic segmentation.}
Semantic segmentation is a task of pixel by pixel object labeling. 
The Fully Convolutional Networks (FCN) \cite{Long15} has introduced Deep Convolutional Neural Network (CNNs) to semantic segmentation.
Following the FCN work, most recent semantic segmentation models are based on CNNs. 
The CNNs architecture for semantic segmentation has been improved using Dilated Convolution \cite{Yu16}, 
ResNets \cite{He16}, \cite{Chen16}, \cite{Wu16}, or DenseNets \cite{Huang17}, \cite{Jegou17}. Deconvolution \cite{Noh15} and Unpooling \cite{Badrinarayanan15} advanced upsampling techniques, 
and the multi-scale down/upsampling has been introduced \cite{Zhao17}. 
Postproccessing techniques such as Conditional Random Fields are also used to improve performance \cite{Chen15}, \cite{Zheng15}. 
Datasets for semantic segmentation such as PASCAL VOC \cite{Everingham15}, MS-COCO \cite{Lin2014}, Cityscapes \cite{Cordts16} and SUN RGB-D \cite{Song15} 
have been proposed. PASCAL VOC contains nearly 10,000 images annotated with pixel by pixel 20 classes,
and MS COCO over 200,000 images annotated with pixel by pixel 80 classes. Cityscapes contains 5,000 images of 
urban driving scenes annotated with pixel by pixel 19 classes. SUN RGB-D contains over 10,000 RGB and Depth image pairs.
The RGB-D image pairs are captured in various indoor scenes and annotated with pixel by pixel 37 classes. According to \cite{Cordts16},
annotating and quality control required more than 90 minutes for a single image. 
Due to the high cost of collecting images and annotating pixel by pixel, some computer graphics datasets for 
computer vision tasks have been introduced.

\paragraph{Computer graphics datasets for training.}
There are object repositories such as ModelNet \cite{Zhirong15} and ShapeNet \cite{Chang15} for 3D shape recognition or reconstruction task. 
Since object repositories cannot be directly used for scene understanding, 
computer graphics datasets have been proposed. SceneNet \cite{Handa16} contains 10K annotated frames without photo-realistic rendering.
SceneNet RGBD \cite{Mccormac17} is an extension of \cite{Handa16} which contains 5M annotated frames with photo-realistic rendering.
Objects are randomly allocated and frames are obtained as sequential video. 
There is a trade-off between rendering time and quality of rendered images. 
In SceneNet RGBD case, each rendering took 2 - 3 sec using Opposite Renderer \cite{Pedersen13}. SUN CG \cite{Song16}, \cite{Zhang16} contains 
400K annotated frames with photo-realistic rendering. 
Objects are manually allocated and frames are obtained from randomly sampled still camera images. 
In SUN CG case, each rendering took about 30 sec using Mitsuba \cite{Jakob10}. 
In Addition, datasets generated by using video game technologies are proposed \cite{Qiu16}, \cite{Richter17}. 
These computer graphics datasets have larger scale than real data and they can be easily increased in the future. 
However, models trained on computer graphics domain data cannot perform well on real domain data due to domain shift.

\paragraph{Domain Adaptation.}
Domain adaptation try to overcome domain shift. 
Recent work based on deep neural network aims to obtain domain invariant features by
end-to-end training \cite{Tzeng15}, \cite{Ganin15}, \cite{Ganin16}, \cite{LongM15}, \cite{LongM16}. 
Other approaches use an adversarial training technique \cite{Bousmalis17}, \cite{Liu16}, \cite{Tzeng17} 
which has been introduced for Generative Adversarial Network \cite{Goodfellow14}. 
They mostly focus on the classification task which has domain shift between commercial images and 
real photographs \cite{Saenko10} or computer graphics data and real data \cite{Sun14}. 
There are very few approaches for semantic segmentation but it has a large impact on the task due to labeling cost.
FCN in the wild \cite{Hoffman16} proposes global alignment by domain adversarial training and local alignment by constraining output.
Curriculum domain adaptation \cite{Zhang17} proposes curriculum learning which solves the easy task first and 
uses them to regularize. 
Domain adaptation with GANs \cite{Sankaranarayanan17} proposed generative models based on CycleGAN \cite{Zhu17} to align 
source and target distributions in the feature space. 
They all evaluate performance on the dataset of driving scene parsing. 
Our approach aims for indoor scene parsing which has domain shift between computer graphics data and real data.
There is another work of adversarial training \cite{Kanazawa17}.
Their method reconstructs 3D mesh of human body from 2D image by factorized adversarial training for each joint location. 
In contrast, our approach factorizes for each object class.

% Head 1
\section{Approach}
Our approach is based on CNNs and adversarial training with discriminators.
First, we present ideas and details of our class-wise adaptation in section 3.1.
Then, we introduce the usage of depth modal and synthetic noise in section 3.2.
Finally, the architecture of neural networks for semantic segmentation and discriminator are described in section 3.3. 

\subsection{Class-Wise Adaptation}
In adversarial training for domain adaptation, the discriminator and the feature extractor are alternately trained.
The discriminator is trained to distinguish whether the feature has been obtained from source domain or target domain, 
and the feature extractor is trained to deceive the discriminator. 
As a result, the feature extractor can extract domain invariant features.
However in semantic segmentation, since the dimension of the features tends to be large, the task becomes more difficult.
In addition, between computer graphics data and real data, the intensity of adaptation should be changed for each object class. 
For instance, a plastic shelf does not need to be adapted as much, but a glass may need to be adapted.
In order to simplify the task and control the intensity of adaptation, we prepare the same number of discriminators as classes and train individually.

\begin{figure}
\includegraphics[keepaspectratio=true,width=\columnwidth]{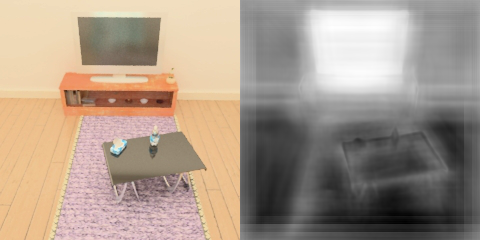}
\caption{The left shows the input rgb image and the right shows normalized features of TV.
 The place of TV tends to bright in the image.}
\label{fig:features_rgbtv}
\end{figure}

To split the features for each class, it is necessary to complete the feature extraction for semantic segmentation.
Therefore, we define the features as the last convolution layer of the semantic segmentation networks.
The channel of the features have the same dimension number as the class number.
For instance, figure~\ref{fig:features_rgbtv} shows the normalized features that are assigned to the TV class.
The location and shape of the TV is obtained.

Furthermore, our discriminators minimize losses for each pixel, not for the whole input.
The accuracy of semantic segmentation is decreased by local domain shift such as location of objects or contours of objects.
Pixel-wise loss which uses local information for domain adaptation helps for local domain shift.

\begin{figure}
\includegraphics[keepaspectratio=true,width=\columnwidth]{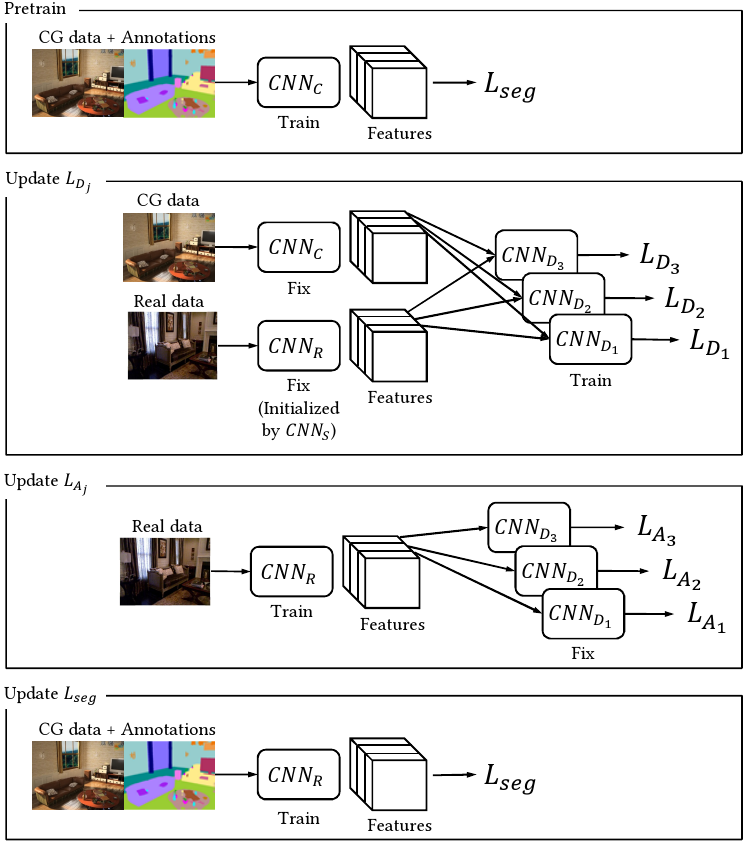}
\caption{The training procedure of class-wise adversarial adaptation. 
 This figure shows an example of 3 classes for simplicity. 
 $CNN_{C}$ and $CNN_{R}$ are the models for semantic segmentation. 
 ${CNN_{D}}_{j}$ is a discriminator of $j^{th}$ class. 
 Features are obtained from the last convolution layer of the models. 
 First, $CNN_{C}$ is pre-trained with only computer graphics data. 
 Then, $CNN_{R}$ and ${CNN_{D}}_{j}$ are iteratively updated by ${L_{D}}_{j}$, ${L_{A}}_{j}$ and $L_{seg}$.}
\label{fig:adversarialtraining}
\end{figure}

Figure~\ref{fig:adversarialtraining} shows the training procedure. 
First, the $CNN_{C}$ is pre-trained by $L_{seg}$ which is the standard supervised semantic segmentation loss (2D Cross Entropy) 
with only computer graphics data. Then, the $CNN_{R}$ is initialized by $CNN_{C}$.
Let ${L_{D}}_{j}$ be the domain loss of $j^{th}$ class 
and $p_{i}$ be the probability distribution for $i^{th}$ domain 
and $d_{i}$ be the one hot vector of ground truth labels
and $k$ be the number of classes for semantic segmentation,
and H be the height and W be the width of the output.
\begin{equation} 
 {L_{D}}_{j} = -\frac{1}{HW} \sum_{y=0}^H \sum_{x=0}^W \sum_{i=0}^1 d_{i} \log p_{i} (x, y, j)
\end{equation}
In the phase of updating ${L_{D}}_{j}$, computer graphics data are inputted to $CNN_{C}$ and real data are inputted to $CNN_{R}$ 
to obtain the features.
The features are split along the channel and inputted to the discriminator ${CNN_{D}}_{j}$ for each class ${j}$.
The weights of $CNN_{C}$ and $CNN_{R}$ are fixed and ${CNN_{D}}_{j}$ are iteratively updated by ${L_{D}}_{j}$.
${CNN_{D}}_{j}$ determines whether a given features have obtained from computer graphics data or real data.
Domain labels are automatically assigned by data sources.
For example, if the computer graphics data are inputted, the domain labels are assigned 0 and 
if the real data are inputted, the domain labels are assigned 1.
No annotated image of semantic segmentation is required in this phase. 
Let ${L_{A}}_{j}$ be the adversarial loss of $j^{th}$ class.
\begin{equation}
 {L_{A}}_{j} = -\frac{1}{HW} \sum_{y=0}^H \sum_{x=0}^W \sum_{i=0}^1 (1 - d_{i}) \log p_{i} (x, y, j)
\end{equation}
In the phase of updating ${L_{A}}_{j}$, real data are inputted to $CNN_{R}$ to obtain the features.
The features are split along the channel and inputted to the discriminator ${CNN_{D}}_{j}$ for each class ${j}$.
The weights of ${CNN_{D}}_{j}$ are fixed and $CNN_{R}$ is iteratively updated by ${L_{A}}_{j}$.
Domain labels are inversed from the phase of updating ${L_{D}}_{j}$.
For example, if the computer graphics domain is assigned 0 and the real domain is assigned 1 in the phase of updating ${L_{D}}_{j}$,
the computer graphics domain is assigned 0 in this phase.
No annotated image of semantic segmentation is required in this phase.
By updating the adversarial loss, $CNN_{R}$ is trained to deceive ${CNN_{D}}_{j}$.
$CNN_{R}$ is also updated by $L_{seg}$ with computer graphics data.
Finally, $CNN_{R}$ and ${CNN_{D}}_{j}$ are iteratively updated by ${L_{D}}_{j}$, ${L_{A}}_{j}$ and $L_{seg}$.

\subsection{Depth and Noise}
Computer graphics can generate various modal for training inputs.
Our experiments show that using depth modal increases accuracy.
The depth image is concatenated with the rgb image on the fourth channel.
At the inference phase, the depth image obtained by a depth sensor often lacks data. 
Therefore, we fill holes of depth image by using image inpainting algorithms \cite{Bertalmio01}.

In general, computer graphics generate smooth images.
However, the images taken by actual cameras have various noises.
In order to bring rendered images closer to real images, we add synthetic noises to rendered images.
Our synthetic noises are based on 2D image processing. 
Gaussian noise is added to the images as random pixel values according to the normal distribution. 
Salt and pepper noise randomly changes the pixels in the image to white or black. 
Gaussian blur and bilateral filter smooth images with kernel size of (9, 9).
These noises are randomly applied for each input (including depth images) when training.

\subsection{Architecture of Neural Networks}
Our CNNs architecture for semantic segmentation is based on DenseNets \cite{Huang17} and PSPNets \cite{Zhao17}. 
DenseNets is built from dense blocks. 
In each layer of dense block, feature maps of all preceding layers are concatenated. 
Therefore, it is easy to reuse the information from previous feature maps. 
The architecture fits for semantic segmentation where skip connections play an important role.
Let $H_{l}$ be the function of a layer and $z_{l}$ be an input of $l^{th}$ layer:
\begin{equation}
z_{l} = H_{l}([z_{0}, z_{1}, . . . , z_{l-1}])
\end{equation}
where $[z_{0}, z_{1}, . . . , z_{l-1}]$ refers to the concatenation of the
feature-maps produced in layers $0, . . . , l-1$.
Let $M_{l}$ be the number of output feature maps and $r$ be the growth rate at $l^{th}$ layer in a denseblock.
$M_{0}$ is determined by the output map size of the layer before the denseblock.
\begin{equation}
M_{l}=M_{0}+rl
\end{equation}
The dilated convolution also can integrate information of multiple receptive fields.
The number of dilation is designed to become larger as the layer in the dense block gets deeper when downsampling phase.
On the other hand, the number of dilation becomes smaller as the layer in the dense block gets deeper when upsampling phase.
Table~\ref{tab:semseg} shows our CNNs architecture for semantic segmentation. 
We define the Dilated Dense as a stack of dense layers with dilation.
Each dense layer has two batchnormalizations \cite{Ioffe15}, two ELU activation \cite{ClevertUH15} and two convolutions (kernel 1$\times$1 and 3$\times$3).
PSPNets have a pyramid pool module. 
To improve performance of semantic segmentation, it is important to understand the contextual relationship and to integrate information from various size of receptive fields. 
For example, the contextual relationship means that a bottle is likely to be on a table. 
Pyramid pool module helps to understand the contextual relationship.
We define the Pyramid Pool as five different scales of average pooling layers. 
After the Pyramid Pool, each output stream is concatenated. 
In addition, the strided convolution is used for the Upsample and the bilinear upsampling is used for the Downsample. 
The input size is 240$\times$240, and the number of channels depends on modal.
In the case of using rgb and depth, they are concatenated and the number of channels becomes four.

We also use the Pyramid Pool for discriminators.
Table~\ref{tab:discriminator} shows our CNNs architecture for discriminators. 
Before the first convolution, ReLU activation \cite{Nair10} is applied for the features that are obtained from the last layer of semantic segmentation.

% Table
\begin{table}
\caption{CNNs architecture for semantic segmentation. 
 The architecture is built with dense blocks, the pyramid pool module, dilated convolutions. We call it DensePyramid. 
 The total number of convolutions is 84.}
\label{tab:semseg}
\begin{minipage}{\columnwidth}
\begin{center}
\begin{tabular}{lcc}
  \toprule
  Block type & Output shape & No. of convs\\
  \hline
  7$\times$7 Conv & 64$\times$240$\times$240 & 1\\
  Average Pool & 64$\times$120$\times$120 & 0\\
  Dilated Dense (3 layers) & 256$\times$120$\times$120 & 6\\
  Downsample & 128$\times$60$\times$60 & 1\\
  Dilated Dense (6 layers) & 512$\times$60$\times$60 & 12\\
  Downsample & 256$\times$30$\times$30 & 1\\
  Dilated Dense (9 layers) & 832$\times$30$\times$30 & 18\\
  Pyramid Pool & 832$\times$30$\times$30 & 5\\
  Dilated Dense (9 layers) & 1088$\times$30$\times$30 & 18\\
  Upsample & 272$\times$60$\times$60 & 1\\
  Dilated Dense (6 layers) & 656$\times$60$\times$60 & 12\\
  Upsample & 164$\times$120$\times$120 & 1\\
  Dilated Dense (3 layers) & 356$\times$120$\times$120 & 6\\
  Bilinear upsampling  & 356$\times$240$\times$240 & 0\\
  3$\times$3 Conv & 256$\times$240$\times$240 & 1\\
  1$\times$1 Conv & 38$\times$240$\times$240 & 1\\
  \bottomrule
\end{tabular}
\end{center}
\end{minipage}
\end{table}

\begin{table}
\caption{CNNs architecture for discriminators. 
 Our discriminator also has the pyramid pool module. 
 After the final convolution layer, the ReLU function is applied as the activation function. 
 We prepare the same number of discriminators as classes for class-wise adaptation.}
\label{tab:discriminator}
\begin{minipage}{\columnwidth}
\begin{center}
\begin{tabular}{lcc}
  \toprule
  Block type & Output shape & No. of convs\\
  \hline
  7$\times$7 Conv & 16$\times$240$\times$240 & 1\\
  Average Pool & 16$\times$120$\times$120 & 0\\
  7$\times$7 Conv & 16$\times$120$\times$120 & 1\\
  Average Pool & 16$\times$60$\times$60 & 0\\
  Pyramid Pool & 16$\times$60$\times$60 & 4\\
  Bilinear upsampling  & 16$\times$240$\times$240 & 0\\
  1$\times$1 Conv & 2$\times$240$\times$240 & 1\\
  \bottomrule
\end{tabular}
\end{center}
\end{minipage}
\end{table}

% Head 1
\section{Computer Graphics Dataset}
We have created 16 computer graphics rooms with different styles such as European, American and Asian. 
Each room has a different size, floor plan and objects.
Unlike SUN CG \cite{Song16} etc., we prepared many kinds of props and put them on the rooms to bring closer to the actual rooms.
Some classes for semantic segmentation have been added (plant, glass, cup, bottle, controller, dish, vase and clock) compared to SUN CG. 
We have to capture actual rooms to implement the mixed-reality application.  
In such applications, the interactions with several props are required, so our dataset is beneficial.
We used Maya for modeling and V-Ray for rendering. 
Maya is a toolset to generate 3D computer graphics assets. 
V-ray is a renderer that provides photo-realistic rendering.
All objects in rooms (excluding floor, door, wall and ceiling) are animated in MAYA. 
Furniture in each room are reallocated 5 times and perturbed by 20 frames at that time. 
Every room has 4 types of lighting conditions (morning, day, night with floor light and night with ceiling light).
Therefore, a total of 400 variations of frames can be obtained for each room. 
In addition, the 4 types of materials change repeatedly every frame. 
As a result, various images can be rendered just by placing fixed cameras.

\begin{figure}
\includegraphics[keepaspectratio=true,width=\columnwidth]{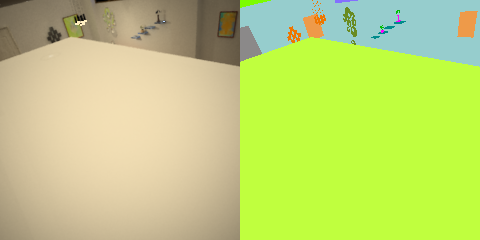}
\caption{The bad annotated image for training semantic segmentation. 
 The left shows the rendered rgb image and the right shows the annotated. 
 The field of view is almost filled with the bookshelf. 
 The CNNs may train beige color as bookshelf.}
\label{fig:badcase}
\end{figure}

In semantic segmentation, the model is trained to understand contextual relationship.
Thus, multiple objects should be included simultaneously in a single image as training data.
As a bad case, if the whole image is filled with single color, the model may train the color as a specific object.
Such a case is shown in Figure~\ref{fig:badcase}.
Randomly placed cameras may generate such data.
In our experiments, we first placed cameras randomly in rooms, but the model did not train well.  
Hence, we placed cameras on the circumference and each face to the center of the rooms.
There are 16 cameras in each room on the circumference at a height of 90 cm and 180 cm.
Our computer graphics can generate 102,400 different images (16 rooms, 400 frames and 16 cameras).
In this work, 20K images which include rgb images, depth images, surface normal images and annotated images were rendered at 480$\times$640.
At the training phase, we use data at 240$\times$240 and augment data by randomly cropping, changing gamma, and flipping etc.

\begin{figure}
\includegraphics[keepaspectratio=true,width=\columnwidth]{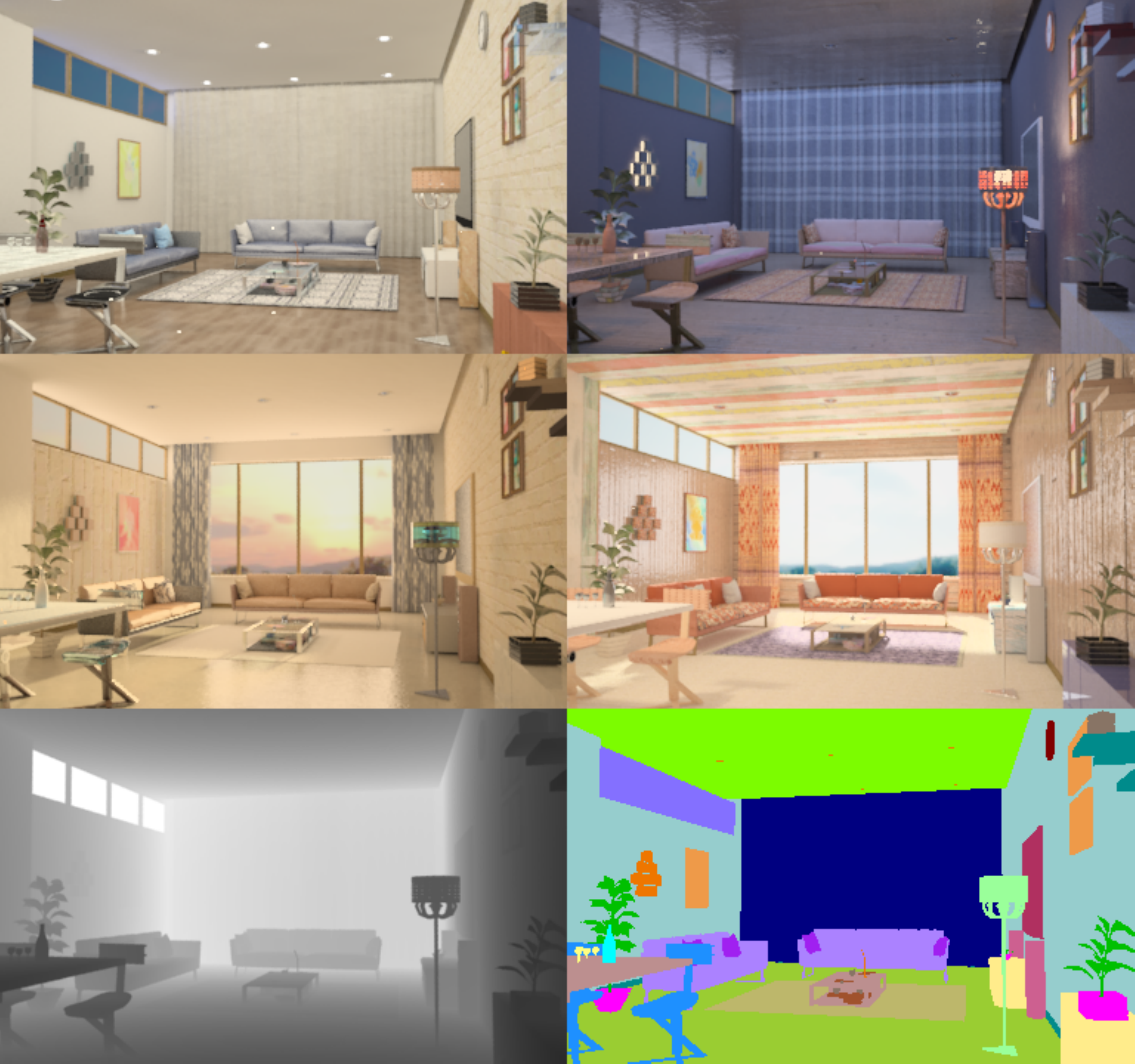}
\caption{Top left shows the rendered rgb image at night with ceiling light, 
 top right shows the rendered rgb image at night with floor light, 
 middle left shows the rendered rgb image at evening and 
 middle right shows the rendered rgb image at morning. 
 Each rendered rgb image has a different materials of furniture. 
 Bottom left shows the rendered depth image and bottom right shows the rendered annotated image. 
 The color of the annotated image means the object class.}
\label{fig:render}
\end{figure}

\begin{figure*}
\includegraphics[keepaspectratio=true,width=\textwidth]{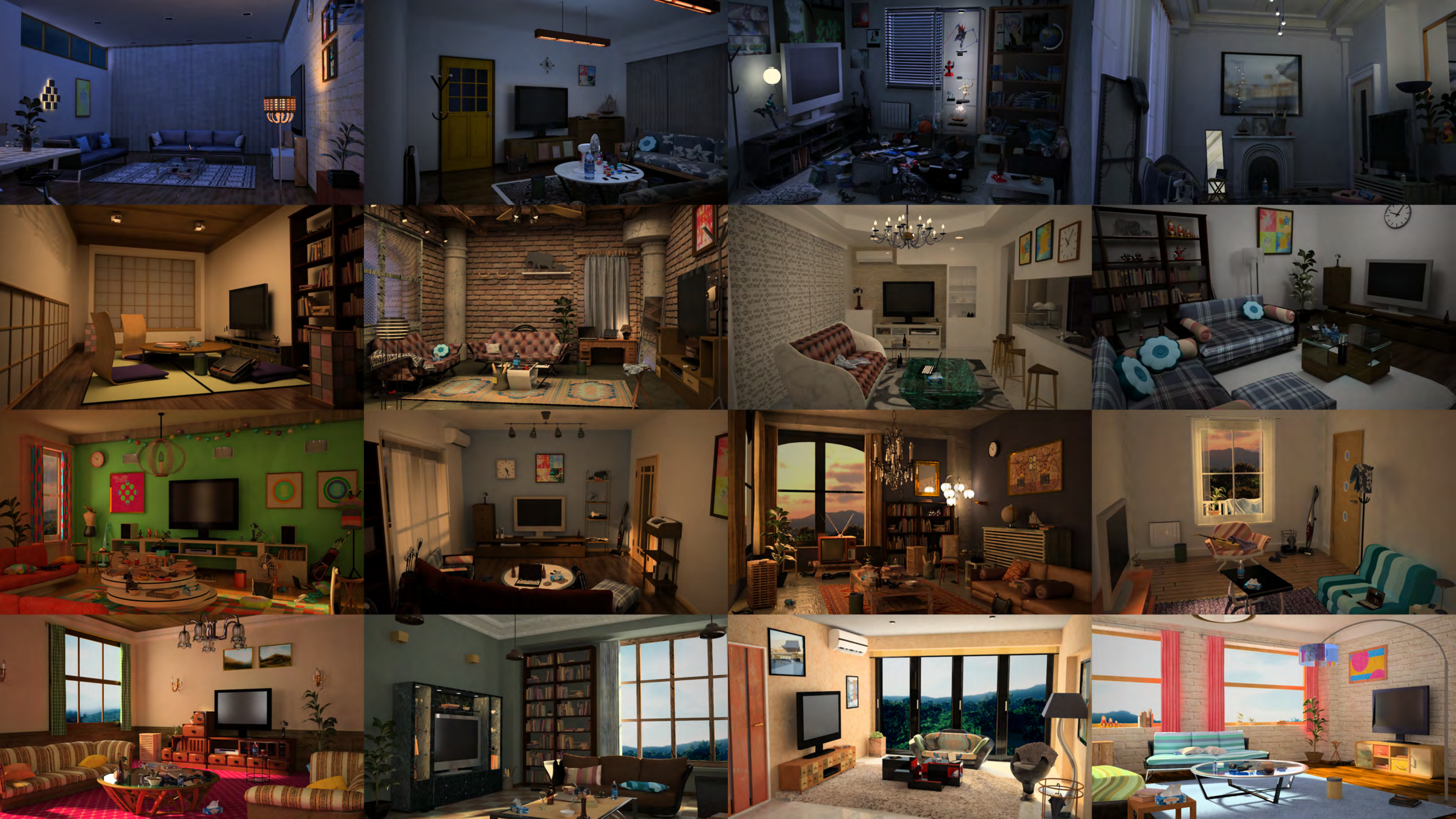}
\caption{The rendered images of all 16 rooms with different conditions. 
 $1^{st}$ row shows night with floor light, $2^{nd}$ row shows night with ceiling light, 
 $3^{rd}$ row shows evening and $4^{th}$ row shows morning. 
 Each room has different style, floor plan, furniture and density of objects.}
\label{fig:renderall}
\end{figure*}

\begin{figure*}
\includegraphics[keepaspectratio=true,width=\textwidth]{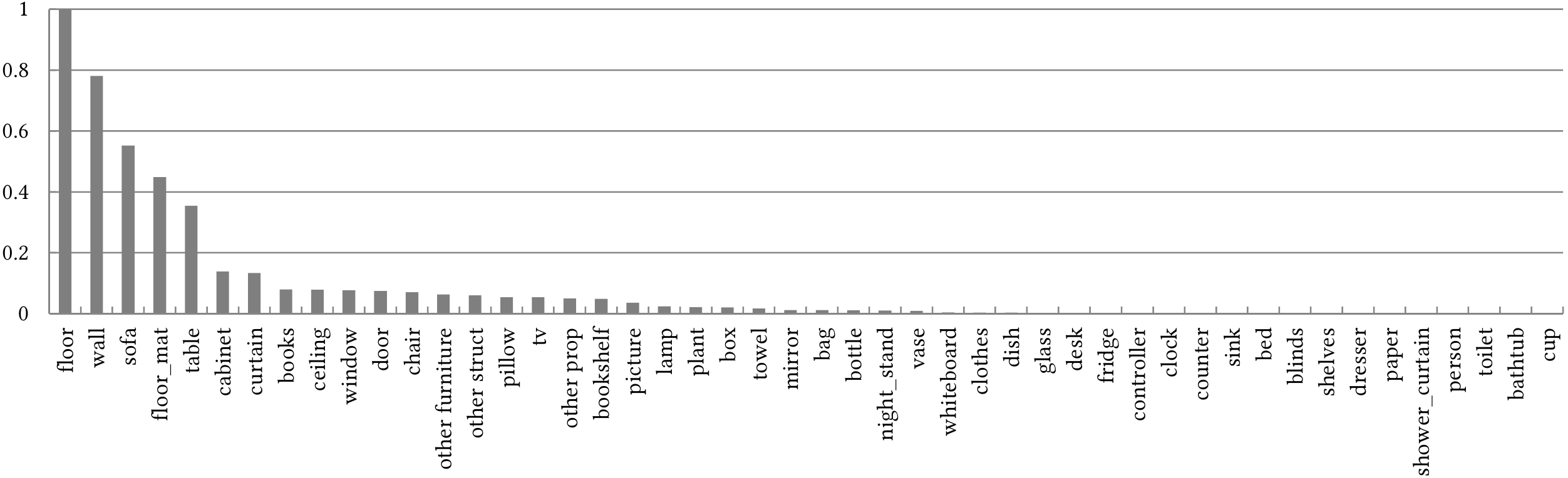}
\caption{The statistics of rendered annotated images of 20K. 
 The ratio with floor labels is shown. 
 Since we have created living room scenes, the items such as bed, shower curtain, toilet and bathtub are zero labels.}
\label{fig:statistics}
\end{figure*}

Rendering took about 90 seconds for a single image. 
The CPU used for rendering is Intel Core i7-6800K. 
We show rendered images in Figure~\ref{fig:render}. 
The rendered images of other rooms are shown in Figure~\ref{fig:renderall} 
and the statistics of annotated images are shown in Figure~\ref{fig:statistics}

% Figure

% Figure

% Head 1
\section{Results}
In this work, we use our computer graphics dataset for the source domain and SUN RGB-D for the target domain.
SUN RGB-D dataset includes various indoor images of rgb and depth, so it is optimal for our task. 
First, we describe training conditions, the evaluation method and comparison with existing method in section 5.1. 
Then, we show the effect of depth and noise in 5.2. 
Finally, we present the effect of class-wise adaptation in 5.3.

% Head 2
\subsection{Implementation Details}
The model for semantic segmentation is trained for 96K iterations with a batch size of 40.
The solver is Adam \cite{Gopalan11} with alpha = 1e-4, beta 1 = 0.9, and beta 2 = 0.999. 
The number of classes is 38, which is the same as SUN RGB-D, classes not used are assigned zero and not trained. 
We evaluate the validation data of SUN RGB-D.
We report the evaluation results using pixel accuracy, mean pixel accuracy, and mean intersection over union \cite{Long15}, \cite{Garcia17}.
Let $n_{ij}$ be the amount of pixels of class $i$ inferred to belong to class $j$ and $k$ be the number of classes.
Pixel Accuracy (PA): the ratio between correctly predicted pixels and the total number of them.
\begin{equation}
PA=\frac{\displaystyle\sum_{i=0}^k n_{ii}}{\displaystyle\sum_{i=0}^k \displaystyle\sum_{j=0}^k n_{ij}}
\end{equation}
Mean Pixel Accuracy (MPA): the average over the total number of classes of class-wise pixel accuracy.
\begin{equation}
MPA=\frac{1}{k+1} \displaystyle\sum_{i=0}^k \frac{n_{ii}}{\displaystyle\sum_{j=0}^k n_{ij}}
\end{equation}
Mean Intersection over Union (MIoU): the ratio between the intersection and the union. 
These are calculated by ground truth pixels and predicted pixels.
\begin{equation}
MIoU=\frac{1}{k+1} \displaystyle\sum_{i=0}^k \frac{n_{ii}}{\displaystyle\sum_{j=0}^k n_{ij} + \displaystyle\sum_{j=0}^k n_{ji} - n_{ii}}
\end{equation}
Table~\ref{tab:neuralnetworks} shows comparison of neural networks. 
DensePyramid is the architecture which was described in section 3.3.
DensePyramid performs 2.1\% higher in MIoU than SegNet.
The models for domain adaptation are trained for 64K iterations with a batch size of 16. 
The solver is stochastic gradient descent (SGD) with learning rate = 1e-6.

% Table
\begin{table}
\caption{Comparison of neural networks. 
 Models are trained with computer graphics data and evaluated with real data. 
 The input modal is rgb.}
\label{tab:neuralnetworks}
\begin{minipage}{\columnwidth}
\begin{center}
\begin{tabular}{lccc}
  \toprule
  Neural Networks & PA & MPA & MIoU\\
  \hline
  SegNet \cite{Badrinarayanan15} & 0.470 & 0.377 & 0.105\\ 
  DensePyramid & 0.470 & 0.368 & 0.126\\ 
  \bottomrule
\end{tabular}
\end{center}
\end{minipage}
\end{table}

% Head 2
\subsection{Effect of Depth and Noises}
We can generate various modal by computer graphics. 
It is important to know which modal is most effective for domain shift between computer graphics data and real data.
In this work, we compare rgb, depth and rgb + depth.
The models are trained with only computer graphics data and evaluated with real data which are included in SUN RGB-D.
We use DensePyramid as the models. 
The results are shown in Table~\ref{tab:modalitiesandnoise}. 
The model trained with rgb + depth performs 7.1\% higher in PA than the model trained with rgb.
The model trained with only depth also performs higher in PA and MPA, however rgb + depth is the best modal.
The result shows that overfitting to color is avoided by using depth images.

We show the effect of synthetic noises in Table~\ref{tab:modalitiesandnoise}. 
We use rgb + depth modal and synthetic noises are added to both rgb and depth.
Salt and pepper noise, gaussian noise, gaussian blur and bilateral filter are applied (see section 3.2). 
These noises are randomly added to input images at training time. 
The model with noises performs around 1\% higher in PA, MPA and MIoU.
Since real data have noises caused by cameras etc., synthetic noises work effectively in such an environment.

% Table
\begin{table}
\caption{Comparison of depth and noise. 
 In the case of rgb + depth, the depth image is concatenated on the fourth channel of input.}
\label{tab:modalitiesandnoise}
\begin{minipage}{\columnwidth}
\begin{center}
\begin{tabular}{lccc}
  \toprule
  Modal and Noise & PA & MPA & MIoU\\
  \hline
  rgb & 0.470 & 0.368 & 0.126\\ 
  depth & 0.489 & 0.392 & 0.120\\ 
  rgb + depth & 0.543 & 0.436 & 0.137\\ 
  rgb + depth + noise & 0.552 & 0.444 & 0.149\\
  \bottomrule
\end{tabular}
\end{center}
\end{minipage}
\end{table}

% Head 2
\subsection{Effect of Class-wise Adaptation}
\begin{figure*}
\centering
\includegraphics[keepaspectratio=true,width=\textwidth]{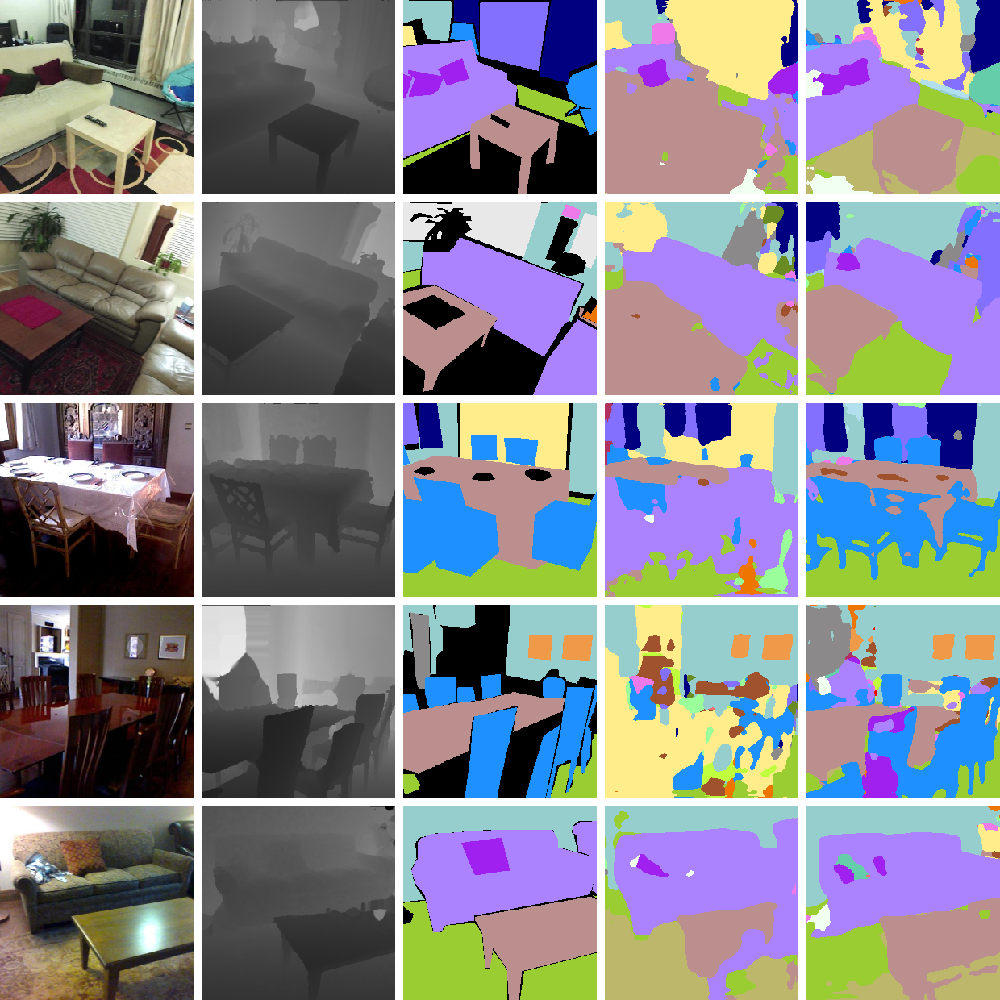}
\includegraphics[keepaspectratio=true,width=\textwidth]{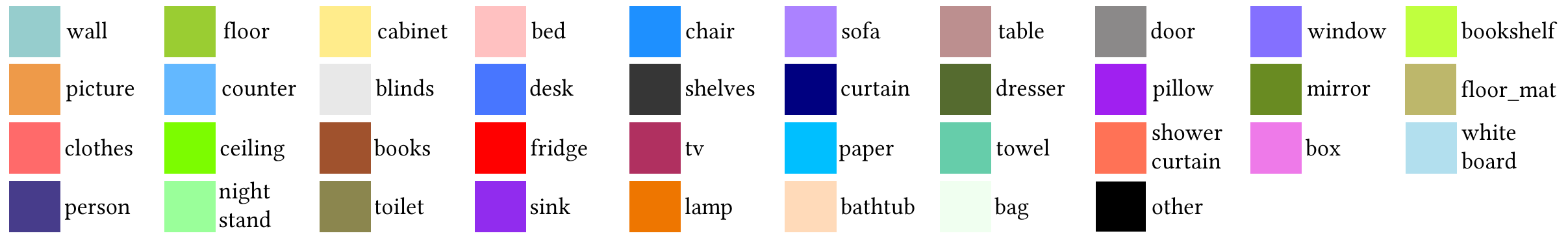}
\caption{Qualitative results of semantic segmentation on the SUN RGB-D validation dataset. 
 From the left column to the right column, input images, smoothed depth image, ground truth, 
 outputs of baseline and outputs of our approach.}
\label{fig:semsegresultsimages}
\end{figure*}

The result of class-wise adaptation is shown in Table~\ref{tab:domainadaptationresults}. 
"Adaptation" in Table~\ref{tab:domainadaptationresults} is a typical method that uses a single discriminator. 
The modal is rgb + depth, and synthetic noises are also added.
The architecture of neural networks is DensePyramid.
DensePyramid and the discriminators are trained with only computer graphics data and evaluated with real data which are included in SUN RGB-D.
Our class-wise adaptation performs 2.3\% higher in PA and 2.2\% higher in MPA, and almost same in MIoU than "no adaptation".
"Adaptation" performs 1\% higher only in MPA, and lower in MIoU.  
The experiment shows that our class-wise adaptation performs better than the typical method.
Some results of sample images are shown in Figure~\ref{fig:semsegresultsimages}. 
The baseline (no depth, no noise and no adaptation) often fails to parse boundaries of objects.
For instance, the table and the floor are confused.
Furthermore, the baseline does not parse chairs well. 
In contrast, our approach parses boundaries of objects well, such as tables, chairs and sofas.
In addition, our approach parses chairs better than the baseline.

% Table
\begin{table}
\caption{Comparison of domain adaptation methods. 
 Adaptation a typical method that uses a single discriminator. 
 }
\label{tab:domainadaptationresults}
\begin{minipage}{\columnwidth}
\begin{center}
\begin{tabular}{lccc}
  \toprule
  Method & PA & MPA & MIoU\\
  \hline
  No Adaptation & 0.552 & 0.444 & 0.149\\
  Adaptation & 0.555 & 0.452 & 0.138\\ 
  Class-wise Adaptation & 0.575 & 0.466 & 0.142\\ 
  \bottomrule
\end{tabular}
\end{center}
\end{minipage}
\end{table}

% Figure

% Head 1
\section{Application}
In this section, we introduce the 3D scene parsing application which is combined semantic segmentation with SLAM.
As a first example, we use elastic fusion as SLAM \cite{Whelan15}, \cite{Whelan16}. 
The elastic fusion can reconstruct dense 3D space in real-time.
The result is shown in Figure~\ref{fig:teaser}.
The left shows actual rgb result and the right shows the parsing result.
The scene is captured as point cloud format, and each color of the parsing result shows each object class.
Since the frame by frame semantic segmentation is unstable, we vote latest several frames to obtain stable results.
Hence, we got the accurate parsing result. 

The whole system including semantic segmentation and SLAM run in 30fps with two NVIDIA Geforce GTX1080s.
It takes less than 1 minute to parse a single room. 
Once the scene is captured and parsed, we can import the real scene into the virtual world and implement a mixed-reality application.
For example, We can teleport to another world with your surroundings by vanishing only walls and change the background.
Or, a real chair is replaced with a computer graphics chair and a virtual character sit on it.
By using head mounted display, the more immersive experience can be realized.

\begin{figure*}
\centering
\includegraphics[keepaspectratio=true,width=\textwidth]{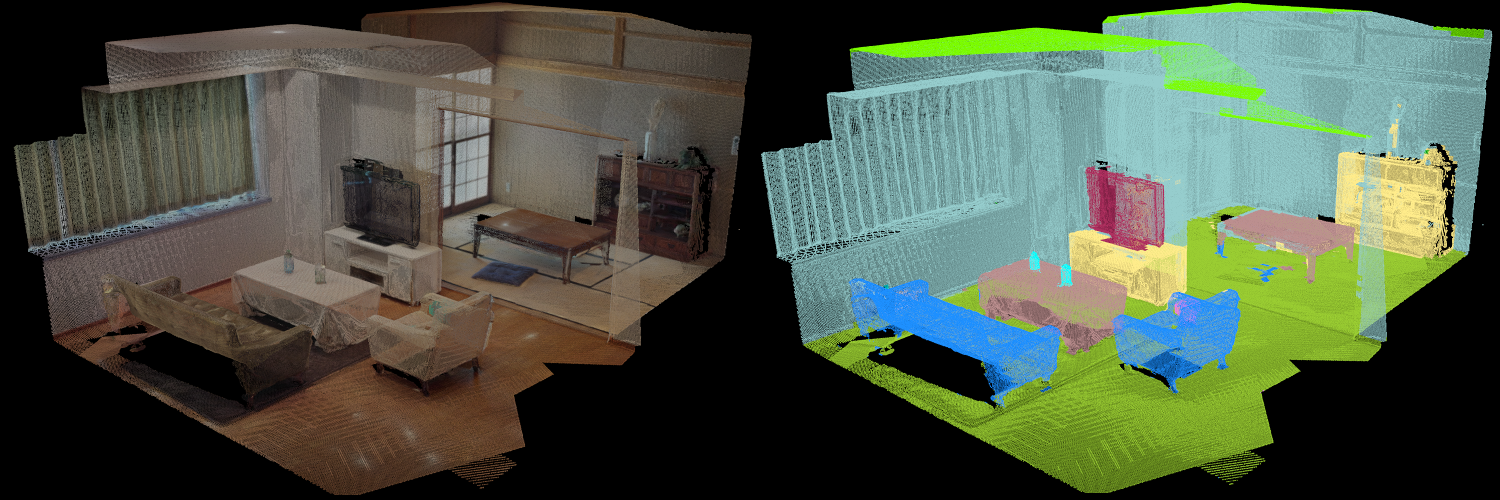}
\caption{The 3D scene parsing result by using Matterport.
The left shows actual rgb image and the right shows the parsing result.
The scene is captured as point cloud format, and each color of the parsing result shows each object class. }
\label{fig:matterport}
\end{figure*}

Another example, we use matterport which is a 3D scanner for rooms. 
Matterport includes cameras and rotates to scan at the fixed position.
We can obtain a whole room or building data by integrating the scanned results at several positions. 
In this work, we use point cloud format.
We obtain the parsing results from virtual cameras, and back-project the results onto the point cloud.
The result by matterport is shown in Figure~\ref{fig:matterport}.
Although it is not in real-time, we can parse the objects with high accuracy by using matterport.

% Figure

% Head 1
\section{Limitations and Future Work}
Since we have not created human computer graphics models, our CNNs do not parse humans.
Various rooms including various human computer graphics models should be prepared to parse humans, 
and we are currently preparing such data.
Since human computer graphics models tend to differ from real humans, the adaptation is assumed to be more difficult.
Therefore, we have to evaluate our approach on humans in the future.

In our data, all TVs are turned off.
It may be necessary to randomly indicate some pictures on the synthetic TV to parse the real TV turned on.
The challenge is how to distinguish between objects in the TV and real objects.
We will tackle these problems as a future work.

The semantic segmentation becomes difficult when only a part of the object appears in the field of view.
There is a possibility to solve that problem using time series data for training.
Therefore, a faster rendering system such as a game engine is required to create time series data.
However, since render quality and render time are a trade-off, 
we have to investigate whether our approach works with poor computer graphics dataset.

% Head 1
\section{Conclusion}
In this paper, we proposed the method that uses only computer graphics datasets to parse the real world 3D scenes.
First, we presented the effectiveness of depth modal and synthetic noise for the domain shift.
The depth image was concatenated with the rgb image on the fourth channel.
Various synthetic noises based on 2D image processing were added to training data.
Then, we developed the class-wise adaptation which can obtain domain invariant features. 
Our class-wise adaptation overcomes the domain shift between computer graphics and the real world without annotated real data.
In addition, we created computer graphics rooms with a lot of props unlike the existing datasets.
Our dataset is beneficial for the actual mixed-reality applications.
Finally, we demonstrated the 3D scene parsing application which is combined semantic segmentation with SLAM.
The latest several frames were voted to obtain stable results.
We got the accurate parsing result in real-time on an actual room.

% Figure

% Figure

\bibliographystyle{ACM-Reference-Format}
\bibliography{main.bib}
\end{document}